\documentclass[10pt,twocolumn,letterpaper]{article}

\usepackage[pagenumbers]{wacv} % To force page numbers, e.g. for an arXiv version

\usepackage{graphicx}
\usepackage{amsmath}
\usepackage{amssymb}
\usepackage{booktabs}
\usepackage{multirow}
\usepackage[dvipsnames]{xcolor}

\usepackage[pagebackref,breaklinks,colorlinks]{hyperref}
\usepackage{amssymb}% http://ctan.org/pkg/amssymb
\usepackage{pifont}% http://ctan.org/pkg/pifont

\usepackage[capitalize]{cleveref}
\crefname{section}{Sec.}{Secs.}
\Crefname{section}{Section}{Sections}
\Crefname{table}{Table}{Tables}
\crefname{table}{Tab.}{Tabs.}

\begin{document}

\title{An Immersive Multi-Elevation Multi-Seasonal Dataset for 3D Reconstruction and Visualization}

\author{
Xijun Liu \qquad  Yifan Zhou \qquad  Yuxiang Guo \qquad Rama Chellappa \qquad Cheng Peng \\
Johns Hopkins University\\
{\tt\small \{xliu253, yzhou223, yguo87, rchella4, cpeng26\}@jhu.edu}
}
\twocolumn[{
  \maketitle
  \vspace{-1.1cm} 
  \begin{center}
    \includegraphics[width=0.95\linewidth]{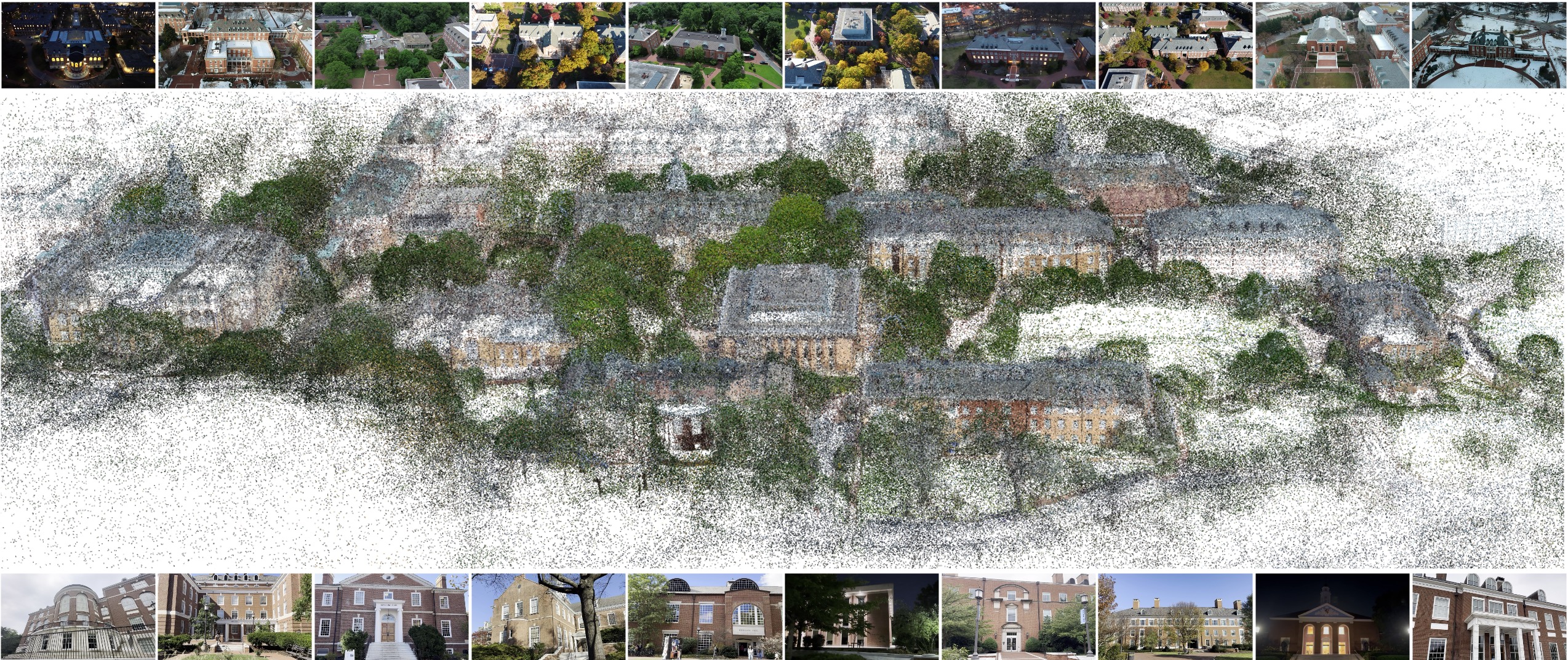}
    \captionof{figure}{A visualization of the reconstructed Johns Hopkins Homewood campus based on our collected imagery over one year.}
    \label{fig:teaser}
  \end{center}
}]

%%%%%%%%% ABSTRACT
\begin{abstract}
   Significant progress has been made in photo-realistic scene reconstruction over recent years. Various disparate efforts have enabled capabilities such as multi-appearance or large-scale modeling; however, there lacks a well-designed dataset that can evaluate the holistic progress of scene reconstruction. We introduce a collection of imagery of the Johns Hopkins Homewood Campus, acquired at different seasons, times of day, in multiple elevations, and across a large scale. We perform a multi-stage calibration process, which efficiently recover camera parameters from phone and drone cameras. This dataset can enable researchers to rigorously explore challenges in unconstrained settings, including effects of inconsistent illumination, reconstruction from large scale and from significantly different perspectives, etc.
\end{abstract}

%%%%%%%%% BODY TEXT
\section{Introduction}
\label{sec:intro}
Three-dimensional (3D) Scene reconstruction is a long-standing research area with extensive applications in robotics, autonomous driving, AR/VR, site modeling, disaster relief planning, etc. Particularly, photorealistic reconstruction can enable immersive experiences in these applications. 
Recent advances in neural rendering, e.g., Neural Radiance Field (NeRF)~\cite{DBLP:conf/eccv/MildenhallSTBRN20} and 3D Gaussian Splatting~\cite{DBLP:journals/corr/abs-2308-04079} have substantially improved our capability in photorealistic novel-view synthesis. Various follow-up work further extend this capability in large scale scenes~\cite{DBLP:conf/cvpr/TurkiRS22, DBLP:conf/cvpr/TancikCYPMSBK22}, to model sites across time~\cite{DBLP:conf/cvpr/Martin-BruallaR21, DBLP:conf/cvpr/ChenZLCFWW22,DBLP:conf/cvpr/YangPW23, DBLP:journals/corr/abs-2407-08447, DBLP:journals/corr/abs-2406-10373, DBLP:conf/eccv/ZhangWWLQW24}, and in various other challenging scenarios~\cite{DBLP:conf/eccv/PengTZWLLC24, DBLP:journals/corr/abs-2404-01133, DBLP:conf/aaai/0008C23}. With these impressive advances, the ability to reconstruction a fully digital world in high fidelity, across time, and at scale may soon become a reality. 
However, these novel methods have been evaluated on scenes that are either collected 1. \textit{at the same time}, 2. \textit{with high overlap}, or 3. \textit{in small scale}. As such, there currently lacks a dataset to realistically evaluate the potential challenges.

Popular datasets in reconstruction, such as Mip-Nerf 360~\cite{DBLP:conf/cvpr/BarronMVSH22} and Tanks and Temples~\cite{DBLP:journals/tog/KnapitschPZK17}, are small scale and limited in appearance variation. 
% and contain few visual ambiguities. 
This controlled environment simplifies camera calibration but does not reflect the complexity of large-scale reconstruction, which may contain many visual ambiguities. Phototourism~\cite{DBLP:journals/tog/SnavelySS06} comprises more complex objects and environments with diverse illuminations and appearances; however, each image is captured at a unique time. As a result, methods~\cite{DBLP:conf/cvpr/Martin-BruallaR21} tested on these datasets requires access to \textit{test-view} images during evaluation to optimize appearance information, potentially compromising the integrity of the test results.
Various datasets~\cite{DBLP:conf/cvpr/TurkiRS22,meuleman2023localrf, DBLP:conf/cvpr/CrandallOSH11, DBLP:conf/iccv/LuYCL0YF23, DBLP:conf/cvpr/TancikCYPMSBK22, DBLP:journals/pami/LiaoXG23, DBLP:conf/cvpr/CaesarBLVLXKPBB20} have been collected to cover a large area, but lack variations in appearance. Furthermore, they are typically collected at the same elevation. Ground-level acquisitions often suffer from occlusions, e.g. induced by plants or walls. This often leads to unconstrained areas of the building, such as the rooftop. Aerial views capture vast areas, but at a reduced resolution to a specific regions. Synthetic datasets~\cite{DBLP:conf/iccv/0002JXX0L023} can offer multi-view imagery of large-scale area under various controlled environments, but they cannot fully reflect the complexities of real-world environments.

We introduce a university-scale, real-world dataset.
This dataset is a comprehensive and expansive outlook of the Johns Hopkins University Homewood Campus, covering area over \underline{80000 $\textrm{m}^2$}. As shown in Fig. 1, the images are captured and selected meticulously around buildings within the campus, and were photographed across \textit{different seasons}, \textit{distinct times of a day}, and \textit{different altitudes}, over a year. Our dataset overcomes previous shortcomings through the following characteristics: (1) Large-scale and potentially repetitive structures: the architecture of the JHU campus buildings has a uniform style. This can introduce visual ambiguities that challenge image registration, where images with similar textures may be matched together. By offering large-scale scenes with intricate architectural patterns, our dataset stresses the robustness of reconstruction algorithms. (2) Multi-appearance and Multi-view: Photographs taken during different seasons and times of day create extensive variations in building appearance and illumination. For each building, we provide images captured around the structure at multiple time of the day and around the year, ensuring uniform appearances during a single acquisition. This approach yields multi-view datasets that allow a fair evaluation of methods with access to only metadata, such as the timestamp, and not the groundtruth image. (3) Multi-elevation: To provide a comprehensive display of each building’s structure, our dataset includes imagery captured from the ground-level, aerial views at various altitudes. By combining ground-level and multi-altitude acquisitions, we enable algorithms to reconstruct buildings with rich details at every angle.

While commercial products are available for accurate ground GPS positioning, they are much less accessible than phones. Since weather conditions can be fast changing, we elect to use phones to capture most of our data. To achieve accurate camera pose estimation from large-scale collections, 
we employ a multi-stage image registration process. First, we simultaneously register all images of each individual building, incorporating additional constraints into the registration algorithm avoid visual doppelgangers. Next, we establish a stable anchor coordinate system using a subset of images captured in summer from 60m above ground. By aligning each building’s registered cameras to this anchor coordinate system, we integrate all building reconstructions into a unified spatial reference. As a result, we can achieve a coherent, large-scale sparse reconstruction of the Johns Hopkins University Homewood Campus with a reasonable processing time.

% \end{enumerate}
\section{Related Work}
\subsection{Scene Level Reconstruction Datasets}
In scene reconstruction research, the widely used benchmark datasets often focus on single objects ~\cite{DBLP:conf/eccv/MildenhallSTBRN20,DBLP:journals/tog/KnapitschPZK17,DBLP:conf/cvpr/BarronMVSH22} or indoor scenes~\cite{DBLP:conf/wacv/LinCDWQH18}. These datasets are collected in controlled environments, allowing highly accurate camera estimation. However, these datasets do not reflect various practical issues, including changes in appearances.
There have been various attempts~\cite{DBLP:journals/tog/SnavelySS06, DBLP:conf/eccv/TungCCYZWHS24, DBLP:conf/cvpr/0008LLZRZFQ20} to construct outdoor unbounded architecture datasets with diverse lighting and appearance conditions. While these datasets include appearance diversity, they lack multi-view imagery with a consistent appearance. Consequently, algorithms~\cite{DBLP:conf/cvpr/Martin-BruallaR21, DBLP:conf/cvpr/ChenZLCFWW22,DBLP:conf/cvpr/YangPW23, DBLP:journals/corr/abs-2407-08447, DBLP:journals/corr/abs-2406-10373, DBLP:conf/eccv/ZhangWWLQW24} tested on these datasets require access to test-view images during evaluation to account for unique appearance variation. Such an approach risks compromising evaluation fairness and does not reflect the inference condition in reality. Ideally, a single appearance should be captured across multiple views, such that the held-out test views can be rendered based on time instead of pixel-wise information.

\subsection{Large-scale Reconstruction Datasets}
Large-scale datasets, such as Quad 6K\cite{DBLP:conf/cvpr/CrandallOSH11}, UrbanScene3D\cite{DBLP:conf/cvpr/CrandallOSH11}, Mill-19\cite{DBLP:conf/cvpr/TurkiRS22}, and OMMO\cite{DBLP:conf/iccv/LuYCL0YF23}, have been collected across a large area from the air. However, these datasets are usually restricted to a single altitude.  This limits the level of detail in reconstructed models. Datasets like Block-NeRF\cite{DBLP:conf/cvpr/TancikCYPMSBK22}, KITTI-360\cite{DBLP:journals/pami/LiaoXG23}, and NuScenes\cite{DBLP:conf/cvpr/CaesarBLVLXKPBB20} focus exclusively on street-level imagery, leading to many unobserved regions such as the roof of the buildings. Although large-scale synthetic datasets, like MatrixCity\cite{DBLP:conf/iccv/0002JXX0L023}, provide both ground and aerial views with a control of the environment and precise ground truth, real-world complexities, including environmental factors and real world physics interactions, cannot be fully reflected. 

\section{Data Collection}
To address the limitations of previously collected datasets, we propose a novel collection designed to bridge the gaps in existing benchmarks. As shown in Table~\ref{tab:comparison}, by capturing real-world, large-scale imagery with multi-appearance and multi-elevation coverage, our dataset provides a comprehensive testing ground for evaluating modern reconstruction algorithms. 

\begin{table}[h]
    \centering
    \resizebox{!}{2cm}{
    \scriptsize
     \begin{tabular}{l|cccccc}
    \toprule
       Dataset & \# images & Scale & mA & mV & Elevation\\
   \midrule
        Phototourism\cite{DBLP:journals/tog/SnavelySS06} & 150K & Scene (R) &  \color{OliveGreen}{\ding{51}} & \textcolor{red}{\ding{55}} & Ground\\
        MegaScenes\cite{DBLP:conf/eccv/TungCCYZWHS24} & 2M & Scene (R) & \color{OliveGreen}{\ding{51}} & \textcolor{red}{\ding{55}} & Ground \\
        BlendedMVS\cite{DBLP:conf/cvpr/0008LLZRZFQ20} & 5K & Scene (R+S)& \textcolor{red}{\ding{55}} & \textcolor{red}{\ding{55}} & Ground\\
        UrbanScene3D\cite{DBLP:conf/cvpr/CrandallOSH11} & 128K & Scene (R+S)& \textcolor{red}{\ding{55}} & \textcolor{red}{\ding{55}} & Aerial \\
        Quad 6K\cite{DBLP:conf/cvpr/CrandallOSH11} & 5.1K & Scene (R)& \textcolor{red}{\ding{55}} & \textcolor{red}{\ding{55}} & Aerial \\
        Mill 19\cite{DBLP:conf/cvpr/TurkiRS22} & 3.6K & Scene (R)& \textcolor{red}{\ding{55}} & \textcolor{red}{\ding{55}} & Aerial \\
        OMMO\cite{DBLP:conf/iccv/LuYCL0YF23} & 14.7k & Scene (R)& \color{OliveGreen}{\ding{51}} & \color{OliveGreen}{\ding{51}} & Aerial \\
        Block-NeRF\cite{DBLP:conf/cvpr/TancikCYPMSBK22} & 2.8M & City (R)& \color{OliveGreen}{\ding{51}} & \textcolor{red}{\ding{55}} & Ground \\
        KITTI-360\cite{DBLP:journals/pami/LiaoXG23} & 300K & City (R)& \textcolor{red}{\ding{55}} & \textcolor{red}{\ding{55}} & Ground \\
        NuScenes\cite{DBLP:conf/cvpr/CaesarBLVLXKPBB20} & 1.4M & City (R)& \color{OliveGreen}{\ding{51}} & \textcolor{red}{\ding{55}} & Ground \\
        MatrixCity\cite{DBLP:conf/iccv/0002JXX0L023} & 519K & City (S)& \color{OliveGreen}{\ding{51}} & \color{OliveGreen}{\ding{51}} & Ground+Aerial \\
    \midrule
        Ours & 12.3K & City (R)& \color{OliveGreen}{\ding{51}} & \color{OliveGreen}{\ding{51}} & Ground+Aerial \\
        
    \bottomrule
    \end{tabular}
    }
    \caption{A comparison of existing 3D reconstruction datasets highlighting key properties, including scale, diversity of appearances, and viewpoint variation. Our dataset distinguishes itself through multi-appearance (mA), multi-view (mV), and multi-elevation (mE) coverage, providing a comprehensive testing ground for evaluating modern reconstruction algorithms. }
    \label{tab:comparison}
    \vspace{-0.3cm}
\end{table}

\begin{figure}
    \centering
    \includegraphics[width=1\linewidth]{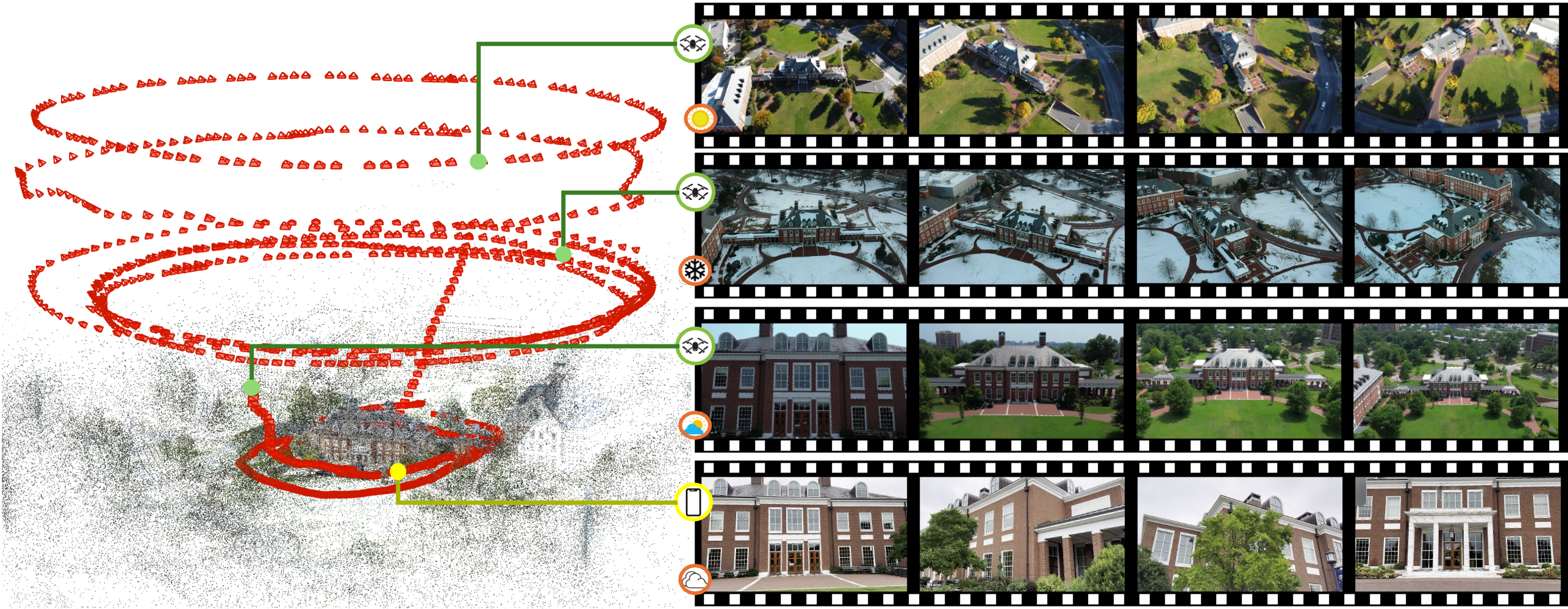}
    \caption{Example of the sparse reconstruction of Mason Hall with the registered camera poses from multi-elevation image sets under various appearance conditions. The displayed configurations include aerial imagery captured in Winter, Summer, and Fall. Ground-level images are taken in Summer and Fall. Two ascending image sequences are also included. These diverse viewpoints and time highlight the comprehensive data collection approach employed in our dataset. }
    \label{fig:collection}
    \vspace{-0.5cm}
\end{figure}

The dataset contains over \textit{12,300 images} on ten adjacent buildings on the Johns Hopkins University Homewood campus, covering an area of approximately \textit{80,000 $\textrm{m}^2$}. As shown in Figure~\ref{fig:collection}, for each building, we collect both aerial and ground-level imagery and systematically vary the appearance configurations. 
In particular, we consider four seasons, weather conditions, such as sunny and cloudy, and times of day, i.e. daytime and nighttime, over the course of one calendar year. The seasonal variations are visually distinct: in Winter, foliage is sparse and often accompanied by snowfall. In Spring and Summer, vegetation growth leads to a significantly greener scene. During Fall, the leaves gradually turn vibrant shades of yellow, orange, and red. In addition to seasonal differences, weather conditions also influence the imagery. On sunny days, strong directional lighting leads to shadows. In contrast, cloudy conditions diffuse the available light, yielding more uniform, ambient illumination. 
We collect a set of multi-view images for each of the mentioned conditions, minimizing the appearance differences within a single collection.

For \textbf{ground-level} imagery, the data collection process involves walking around each building’s perimeter with a handheld smartphone camera, capturing continuous video sequences. We perform a manual inspection of all extracted frames to remove any low-quality images and ensure consecutive frames maintain sufficient overlap to support robust image registration.
In addition, images containing Personally Identifiable Information (PII), such as faces or vehicle license plates, are blurred.

For \textbf{aerial} imagery, we deploy drones equipped with stabilized high-resolution cameras. At each targeted altitude (60m, 100m, and 120m), the drone follows a circular flight trajectory around the building. Drone flights are planned to ensure uniform and complete coverage. 
In addition to circular flights, we also keep the ascending video sequences as the drone moves from ground level to approximately 60m above ground on two sides of each building. These vertically ascending videos can help improve registration between ground-level and aerial imagery.
From these videos, we sample individual frames, applying the same quality control measures as for ground-level data. 

\section{Data Processing}
\begin{figure*}[!htb]
    \setlength{\abovecaptionskip}{3pt}
    \setlength{\tabcolsep}{1pt}
    \begin{tabular}[b]{cccc}
        \begin{subfigure}[b]{.245\linewidth}
            \includegraphics[width=\textwidth]{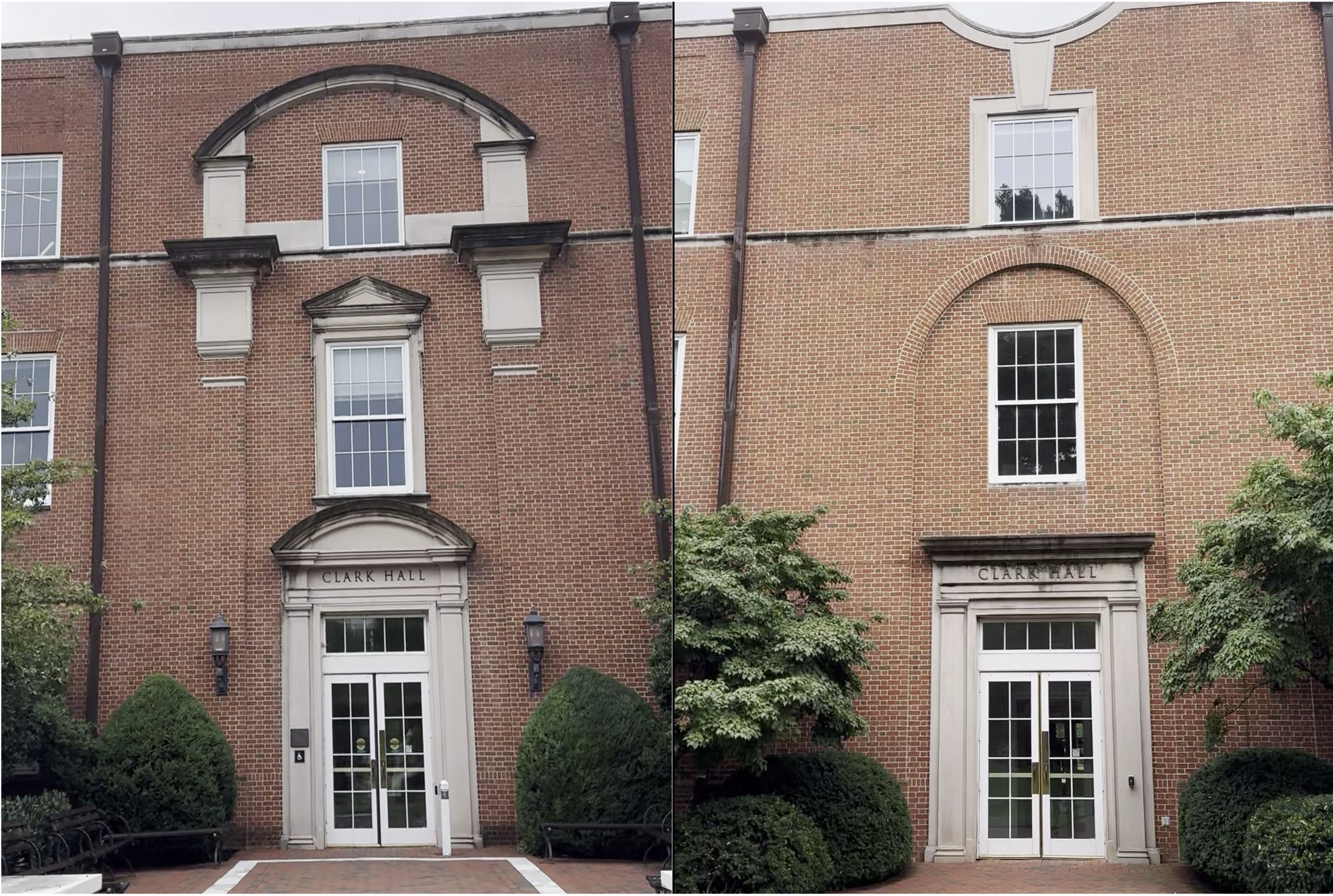}
            \caption{Front vs. back door of Clark Hall}
            \label{front_back}
        \end{subfigure} &
        \begin{subfigure}[b]{.245\linewidth}
            \includegraphics[width=\textwidth]{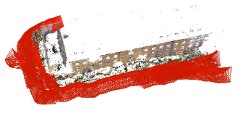}
            \caption{SIFT~\cite{DBLP:journals/ijcv/Lowe04}}
            \label{sift}
        \end{subfigure} &
        \begin{subfigure}[b]{.245\linewidth}
            \includegraphics[width=\textwidth]{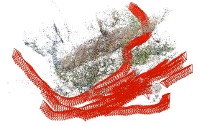}
            \caption{SP+SG~\cite{DBLP:conf/cvpr/SarlinDMR20}}
            \label{spsg}
        \end{subfigure} &
        \begin{subfigure}[b]{.245\linewidth}
            \includegraphics[width=\textwidth]{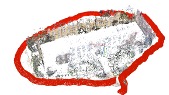}
            \caption{SP+SG with constraint}
            \label{spc}
        \end{subfigure}
    \end{tabular}
    \caption{An example of visual ambiguities, or ``doppelgängers'', observed in our dataset. The front and back of door of Clark Hall is similar, but should not be matched together. Naively using common feature matching algorithms leads to incorrect calibration.}
    \label{fig:method_vis}
\end{figure*}

After acquiring and inspecting the imagery, we apply a multi-stage data processing pipeline to register all images into a single coordinate space. 

\subsection{Large-scale Doppelgänger Mitigation}

Visual ambiguity is a common challenge in calibrating images of buildings, due to their symmetries. This problem, often known as doppelgängers, is also seen in our collected imagery. Without accurate GPS, images that are far apart can get matched erroneously due to similar textures and limited field-of-view. 
As illustrated in Figure~\ref{front_back}, the front door and back door regions of Clark Hall are very similar. Standard feature matching methods, e.g., SIFT~\cite{DBLP:journals/ijcv/Lowe04} and Superglue~\cite{DBLP:conf/cvpr/SarlinDMR20}, either fail to converge or created a heavily overlapped calibration. 

Rather than allowing arbitrary matches across all ground-level images, we imposed a temporal adjacency constraint: each image is only matched with its 10 nearest frames in the video. By confining matches to successive frames, we effectively reduce the search space for feature correspondences and prevented spurious matches. This constraint led to a stable registration of ground-level imagery, even in the presence of repeated architectural motifs. However, registering these images without prior information about image order remains a challenging task and research direction.

\subsection{Multi-elevation imagery integration}
Our dataset spans multiple altitudes, ranging from ground level up to 120m, providing comprehensive and detailed coverage of the building exteriors and rooftops. 
However, drastic altitude and perspective changes pose a distinct challenge for registration. Benchmark registration methods struggled to match the fine-grained features in ground images to drone image. 
To address this issue, we incorporate the ascending image sequences acquired by drones. These sequences start close to ground level and incrementally ascend to higher altitudes, gradually shifting perspectives from ground to air. This incremental transition allows features detected in ground-level images to be traced upward. As such, we can register ground and aerial imagery of together for a single building.
In Table \ref{tab:multielevation_table} we compared the current registration benchmarks on a selected sets of imagery, withholding the ascending image sequences. All methods fail to register cross-view images correctly, prompting further research in this direction.

% \begin{table}
%     \centering
%     \resizebox{!}{2.1cm}{\begin{tabular}{c|cc |c |c |c |c |c}
%     \toprule
%        \multirow{2}{*}{Buildings} & \multicolumn{2}{c|}{\# Images} & \multirow{2}{*}{SIFT} & \multirow{2}{*}{SP+SG}  & \multirow{2}{*}{Loftr} & \multirow{2}{*}{Roma} & \multirow{2}{*}{SP+Roma}\\
%         & G & D &  &  &  & \\
%    \midrule
%         % Ames & &  & 544085 & 725843 & 930723\\
%         Ames &175 & 35 & 175 & 175&175 & 188 & 210\\
%         Clark &171 &33 & 161& 171 & 168 & 180 & 171 \\
%         Garland  &153 & 32&106 & 153 & 102 & 167 & 153\\
%         Gilman & 68 & 48 & 48 & 116 & 48 & 116 & 116 \\
%         Hackerman & 125 & 38 & 122 & 125 & 125 & 119 & 162\\
%         Hodson  &77 & 25& 51 &77  & 63 & 91 & 97\\
%         Latrobe  & 80 & 29& 80 & 80 & 71 & 100 &  109\\
%         Levering  &96 & 43 & 56 & 61 & 78 & 113 & 60 \\
%         Mason  & 143 & 27& 143 & 143 & 143 &160 &170\\
%         Shriver  & 97 & 46 &  75&  97 &97 & 139 &97\\
%     \bottomrule
%     \end{tabular}
%     }
%     \caption{Comparison of the number of images registered by different methods. Here, G represents the number of images collected on the ground, and D denotes those captured by the drone.}
%     \label{tab:multielevation_table}
% \end{table}
\begin{table}
    \centering
    \resizebox{!}{2cm}{\begin{tabular}{c|cc |c c c c c}
    \toprule
       \multirow{2}{*}{Buildings} & \multicolumn{2}{c|}{\# Images} & \multirow{2}{*}{SIFT~\cite{DBLP:journals/ijcv/Lowe04}} & \multirow{2}{*}{SP+SG~\cite{DBLP:conf/cvpr/SarlinDMR20}}  & \multirow{2}{*}{LoFTR~\cite{DBLP:conf/cvpr/SunSWBZ21}} & \multirow{2}{*}{RoMA~\cite{DBLP:conf/cvpr/EdstedtSBWF24}} \\
        & G & D &  &  &  & \\
   \midrule
        % Ames & &  & 544085 & 725843 & 930723\\
        Ames &175 & 35 & 175 & 175&175 & 188 \\
        Clark &171 &33 & 161& 171 & 168 & 180  \\
        Garland  &153 & 32&106 & 153 & 102 & 167 \\
        Gilman & 68 & 48 & 48 & 116 & 48 & 116 \\
        Hackerman & 125 & 38 & 122 & 125 & 125 & 119 \\
        Hodson  &77 & 25& 51 &77  & 63 & 91 \\
        Latrobe  & 80 & 29& 80 & 80 & 71 & 100 \\
        Levering  &96 & 43 & 56 & 61 & 78 & 113 \\
        Mason  & 143 & 27& 143 & 143 & 143 &160\\
        Shriver  & 97 & 46 &  75&  97 &97 & 139\\
    \bottomrule
    \end{tabular}
    }
    \caption{Comparison of the number of images registered by different methods. Here, G represents the number of images collected on the ground, and D denotes those captured by the drone.}
    \label{tab:multielevation_table}
    \vspace{-0.6cm}
\end{table}

\subsection{Global Alignment}

After correctly registering multi-appearance imagery of a single building, we then put all campus building in to a unified coordinate system, which is more efficient than directly registering all available together. 
To accomplish this, we first use a subset of aerial images from every building, captured during summer from an altitude of 60m. Since aerial images have a large perspective, they can be reliably calibrated. This produces a set of camera positions in a campus-wise coordinate system, $C^i_{\textrm{campus}}$. The same aerial camera positions can be expressed in a building-wise coordinate system, $C^i_{\textrm{building}}$, where $i$ indicate the building index.
Then, we perform Procrustes Alignment\cite{gower1975generalized} to align each registration and point cloud for individual buildings with the anchor coordinate system.

A similarity transform comprising a scaling factor \(s \in \mathbb{R}^{+}\), a rotation matrix \(r \in \mathbb{R}^{3 \times 3}\), and a translation vector \(t \in \mathbb{R}^{3}\) are estimated. 
We determine $(s, r, t)$ by solving the following optimization problem:
\begin{align}
(s^{*}, r^{*}, t^{*}) = \arg\min_{s,r,t} \sum_{i}\| s(rC^i_{\textrm{hall}} + t) - C^i_{\textrm{campus}} \|^{2}
\end{align}

Once the optimal $(s^{*}, r^{*}, t^{*})$ is found, it is used to transform all cameras and 3D points to a campus-wise coordinate system.

\section{Conclusion}

We introduce a comprehensive multi-view dataset of the Johns Hopkins University Homewood Campus designed to advance research in high-fidelity camera calibration and reconstruction. By systematically varying seasonal conditions, times of day, weather patterns, and elevations, our dataset encompasses a rich diversity of visual appearances. This breadth enables robust testing and benchmarking of algorithms under challenging real-world scenarios, including visually ambiguity and significant perspective changes.

After collecting the dataset, we outlined a multi-stage data processing pipeline that efficiently calibrates all the images into a single coordinate system. 
Specifically, this is done by 1. imposing temporal constraints to mitigate “Doppelgänger” issues in ground-level imagery, 2. leveraging ascending image sequences to bridge the gap between ground and aerial perspectives, and 3. aligning individual building reconstructions into a unified anchor coordinate system.

Our multi-view calibrated dataset offers a valuable new resource for the computer vision and graphics community. By enabling more rigorous evaluation of 3D reconstruction methods and facilitating comparisons across a wide range of environmental conditions, our work lays the groundwork for advancing state-of-the-art techniques and enhancing their applicability to complex, real-world environments.

%%%%%%%%% REFERENCES
{\small
\bibliographystyle{ieee_fullname}
\bibliography{egbib}
}

\end{document}